\documentclass[runningheads]{llncs}

\usepackage{eccv}

\usepackage{eccvabbrv}

\usepackage{graphicx}
\usepackage{subcaption}
\usepackage{booktabs}
\usepackage{multirow}
\usepackage{xcolor}
\definecolor{first}{HTML}{C5E0B4}
\definecolor{second}{HTML}{E2F0D9}
\definecolor{third}{HTML}{FFF2CC}

\usepackage[accsupp]{axessibility}  % Improves PDF readability for those with disabilities.

\usepackage{hyperref}

\usepackage{orcidlink}

\begin{document}

\title{Proximity-CLIP: Text-Guided Semantic Proximity Learning for Zero-Shot Anomaly Detection} 

\titlerunning{Proximity-CLIP}

\author{Manwen Yang\inst{1,2}\orcidlink{0009-0007-9488-8969} \and
Leqian Ding\inst{2}\orcidlink{0009-0008-8351-1805} \and
Yu Guo\inst{2}\orcidlink{0000-0002-5489-8288}\thanks{Corresponding author: yu.guo@xjtu.edu.cn} 
\and
Fei Wang\inst{2}\orcidlink{0000-0003-3462-8472}}

\authorrunning{M.~Yang et al.}

\institute{School of Software Engineering, Xi'an Jiaotong University \and
 State Key Laboratory of Human-Machine Hybrid Augmented Intelligence, Institute of Artificial Intelligence and Robotics, Xi'an Jiaotong University}

\maketitle

% 摘要
\begin{abstract}

Vision-language models offer a promising approach for zero-shot anomaly detection (ZSAD). However, due to object-centric bias, normal and anomalous text prototypes exhibit a high semantic overlap. While enforcing strict orthogonality between them improves discriminability, mapping highly contiguous visual inputs onto drastically orthogonal prototypes introduces a geometric dilemma, disrupting the pre-trained structural continuity. To address this problem, we propose Proximity-CLIP, a framework that visually calibrates the semantic margin to guide visual adaptation. First, we introduce a visually-calibrated semantic proximity learning mechanism that uses a bounded dynamic regularization to learn an appropriate semantic margin, ensuring discriminative separation while preserving structural alignment. Second, we design an Anomaly Query Module (AQM) driven by these text priors. Using the calibrated anomalous prototype as a semantic query, the AQM actively retrieves localized defect cues from contextual visual patches, mitigating the dilution of subtle anomalies during global pooling. Extensive experiments demonstrate that Proximity-CLIP outperforms current state-of-the-art methods across multiple ZSAD benchmarks with minimal architectural modifications.


\end{abstract}
% 1 introduction
\section{Introduction}
\label{sec:intro}
\begin{figure*}[!t] % figure* 用于双栏排版中的跨栏大图
    \centering
    \includegraphics[width=0.95\textwidth]{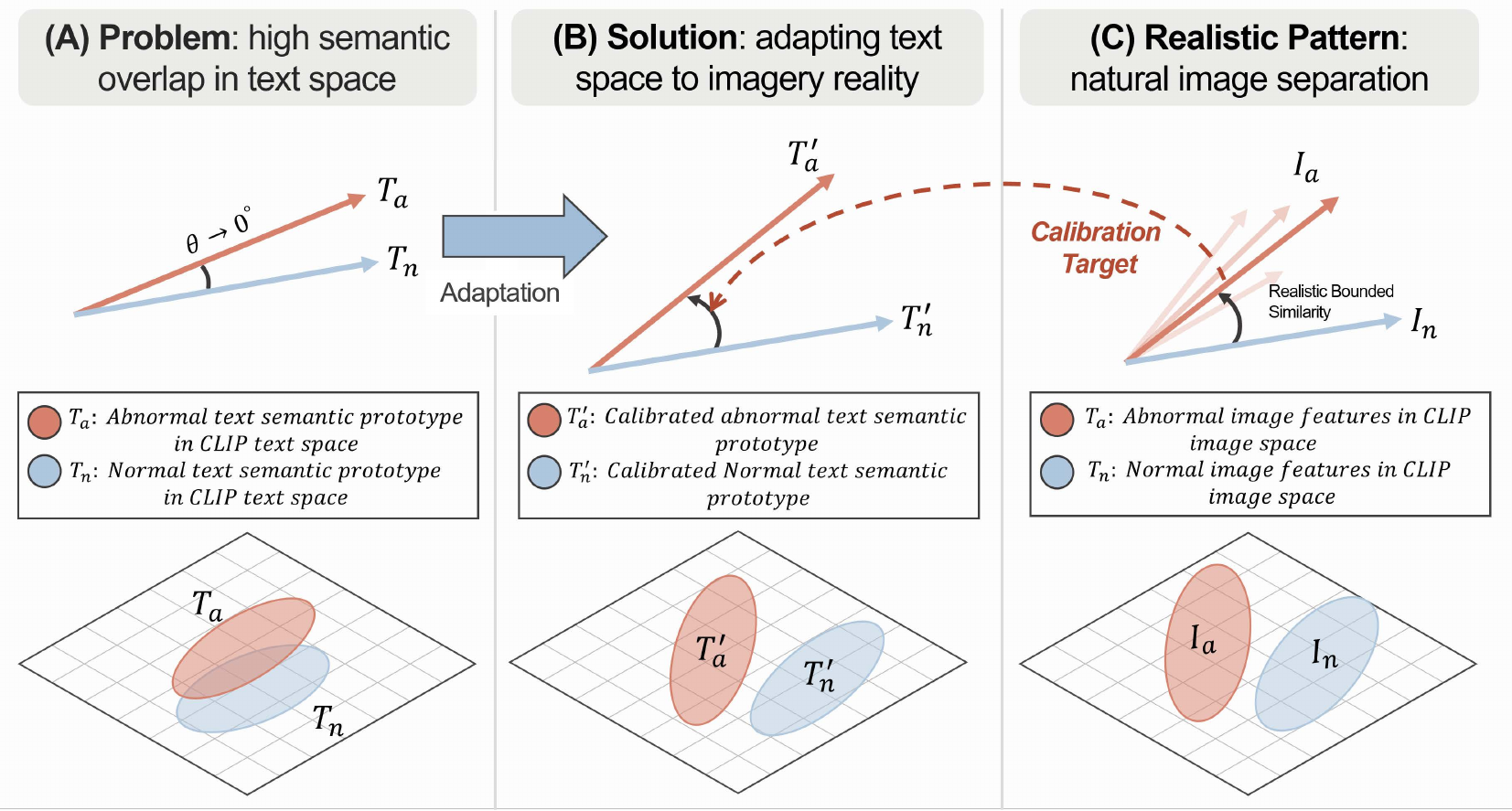} % 略小于全文宽度，看起来更美观
    \caption{\textbf{Visually-calibrated semantic proximity shaping for anomaly detection.} \textbf{(A):} In the pre-trained CLIP space, normal and anomalous text prototypes $\mu_n$ and $\mu_a$ exhibit high semantic overlap due to object-centric bias. \textbf{(B):} Our method explicitly calibrates this proximity, adapting textual distributions to a realistic boundary. \textbf{(C):} This calibrated space mirrors the inherently bounded visual separation between real-world normal and anomalous images $I_n$ and $I_a$.}
    \label{fig:teaser}
\end{figure*}
% ZSAD任务背景与CLIP的局限性
Zero-shot anomaly detection, or ZSAD, localizes anomalous image regions without relying on closed-set normal training samples. Traditional methods that rely on category-specific data for feature reconstruction lack scalability to open-world scenarios. Large-scale vision-language models, such as CLIP, offer a new zero-shot perception approach by aligning image and text representations. However, because these models are pre-trained primarily for instance-level object recognition, they lack the fine-grained sensitivity required for localized structural variations. This deficiency renders the adaptation of CLIP for ZSAD highly challenging.

% 当前PEFT方法的缺陷：离散语义隔离与被动特征对齐 (引入Adapter)
To stimulate the anomaly perception capabilities of CLIP, recent approaches~\cite{ma2025aa,cao2024adaclip,zhou2023anomalyclip} rely on parameter-efficient fine-tuning, which primarily encompasses prompt tuning and adapter tuning, to construct semantic prototypes for normal and abnormal states. Adapter-based methods~\cite{ma2025aa} enforce strict orthogonality between text prototypes to maximize classification margins. While this discrete semantic isolation enhances mathematical discriminability, mapping highly contiguous visual inputs, where anomalies are merely minor localized perturbations on intact backgrounds, onto drastically orthogonal semantic prototypes inevitably introduces a geometric dilemma. This strict regularization disrupts the continuous feature structure of the pre-trained metric space, potentially compromising generalization capabilities on unseen categories.

% 本文提出的Proximity-CLIP框架及阶段一：语义校准
To address this structural dilemma, we propose Proximity-CLIP, a framework that visually calibrates the semantic margin to guide visual adaptation. By modeling an anomaly as a proximate semantic shift structurally bounded by the normal feature distribution(see Fig.~\ref{fig:teaser}), we introduce a coherent two-stage strategy. In the first stage, we propose a visually-calibrated semantic proximity learning mechanism to preserve structural alignment. By formulating the generation of text prototypes as a continuity-constrained optimization problem, we introduce bounded dynamic proximity regularization. Driven entirely by the actual visual variance, this mechanism dynamically learns an appropriate semantic margin to relax rigid angular constraints. Consequently, the model autonomously establishes a calibrated semantic margin that ensures discriminative separation while preserving the pre-trained structural integrity.

% 阶段二：文本先验驱动的主动视觉感知
These semantically calibrated prototypes provide explicit inductive biases for subsequent visual adaptation. In the second stage, we design an Anomaly Query Module (AQM) driven by these text priors. Standard visual adaptation typically processes image patches uniformly, which often causes subtle localized defects to be overshadowed by dominant normal backgrounds during global pooling. To bridge this semantic gap, we introduce a semantic-guided querying mechanism. By using the calibrated anomalous prototype as a driving semantic query, the AQM actively retrieves localized defect cues from the contextual visual patches. This module acts as an explicit top-down spotlight, effectively aggregating fine-grained anomalous responses across multi-level visual hierarchies, thereby shifting the process from passive global matching to active, target-aware anomaly retrieval.

% 核心贡献总结
The main contributions of this paper are summarized as follows:
\begin{itemize}
\item We propose Proximity-CLIP, a framework that reformulates ZSAD through text-guided semantic proximity learning, which addresses the geometric dilemma introduced by strict discrete semantic isolation.
\item We design two synergistic components, including a visually-calibrated semantic proximity learning mechanism that introduces a bounded dynamic regularization to learn an appropriate semantic margin, and an Anomaly Query Module that uses the calibrated anomalous prototype as a semantic query to actively retrieve localized defect cues. 
\item Extensive experiments demonstrate that Proximity-CLIP outperforms current state-of-the-art methods across multiple zero-shot anomaly detection benchmarks while requiring minimal architectural modifications.
\end{itemize}
% 2 related works
\section{Related Work}
\label{sec:related}

\subsection{Traditional Anomaly Detection}
Prior to the integration of vision-language models, the paradigm of visual anomaly detection was dominated by methodologies trained strictly on defect-free samples from specific categories. These traditional approaches are primarily classified into reconstruction-based~\cite{deng2022anomaly,he2024diffusion,lu2023hierarchical,yao2023one,you2022adtr} methods and feature embedding-based~\cite{liu2023simplenet,wang2021student,you2022unified,zhang2023destseg,defard2021padim,gudovskiy2022cflow,kim2023sanflow,zong2018deep,roth2022towards} methods. Reconstruction-based architectures, including autoencoders and generative adversarial networks, operate on the assumption that a network trained exclusively on normal patterns will generate measurable discrepancies when attempting to reconstruct anomalous regions. Conversely, embedding-based approaches, such as PatchCore~\cite{roth2022towards} and PaDiM~\cite{defard2021padim}, project image data into high-dimensional latent spaces, identifying anomalies by calculating the distance between the extracted features of a test sample and the established distribution of normal features. While these conventional models demonstrate significant efficacy in constrained environments, they rely heavily on category-specific data and lack holistic semantic understanding, which severely limits the scalability of such approaches in open-world scenarios.

\subsection{CLIP and Parameter-Efficient Fine-Tuning}
Contrastive Language-Image Pre-training (CLIP)~\cite{radford2021learning} represents a milestone in multimodal learning, utilizing a dual-encoder architecture trained on hundreds of millions of image-text pairs to align visual and textual representations within a shared latent space. 
This robust alignment facilitates powerful zero-shot capabilities without requiring dataset-specific training, demonstrating exceptional performance in open-vocabulary image classification. 
Subsequently, the foundational architecture of this model has been widely adopted to drive diverse downstream tasks~\cite{mokady2021clipcap, patashnik2021styleclip,fang2021clip2video,cho2022fine}, including image captioning~\cite{mokady2021clipcap}, video-text retrieval~\cite{fang2021clip2video}, and image generation and manipulation~\cite{patashnik2021styleclip}. 
Despite these successes in general visual comprehension, the direct application of the pre-trained model to specialized domains often yields suboptimal results due to domain shifts and semantic entanglement. Specifically, because the pre-training objective of CLIP is optimized primarily for global, instance-level object recognition, the learned representations exhibit a strong object-centric bias. Consequently, the base model is insensitive to the fine-grained, localized structural variations, yielding limited performance in dense prediction tasks like semantic segmentation~\cite{wang2022cris, zhou2023zegclip, li2023graphadapter}. This structural deficiency naturally necessitates targeted fine-tuning methodologies to stimulate and adapt the perception capabilities of the pre-trained model for specific downstream applications.

To adapt vision-language models for specialized downstream tasks without inducing catastrophic forgetting, recent surveys highlight the widespread adoption of Parameter-Efficient Fine-Tuning paradigms. These approaches primarily encompass prompt tuning and adapter tuning. Prompt tuning modifies the textual input space by introducing continuous, learnable tokens that optimize the alignment between task-specific semantics and the frozen vision encoder, shifting the paradigm away from manually crafted discrete prompts~\cite{zhou2022learning}. Adapter tuning, alternatively, modifies the internal architecture by inserting lightweight, trainable multi-layer perceptrons or low-rank matrices into the transformer blocks~\cite{gao2024clip,zhang2022tip}. By projecting generic pre-trained features into task-specific subspaces, adapters facilitate direct communication between the vision and language branches, while recent advances further utilize them to eliminate inter-class semantic confusion~\cite{li2025logits} and recalibrate logit biases~\cite{tang2024amu}. These parameter-efficient mechanisms effectively bridge the semantic gap between general pre-training and downstream requirements while strictly preserving the structural integrity of the foundational model.

\subsection{CLIP-Based Anomaly Detection}
Leveraging the generalization capability of CLIP, recent zero-shot anomaly detection frameworks have redefined text-guided visual alignment. Early approaches like WinCLIP~\cite{jeong2023winclip} introduced compositional prompts, while VAND~\cite{chen2023zero} and MVFA-AD~\cite{huang2024adapting} explored multi-level visual feature aggregation to improve pixel-level localization. Advancing this paradigm, subsequent methods such as AnomalyCLIP~\cite{zhou2023anomalyclip} and AdaCLIP~\cite{cao2024adaclip} refined semantic expressiveness through object-agnostic and dynamic prompt learning. More recently, adapter-based techniques like AA-CLIP~\cite{ma2025aa} have sought to enhance discriminability by aligning patch-level visual features with anomaly-aware text anchors. Despite these advancements, fully unleashing CLIP's ZSAD capabilities remains fundamentally challenging. Because physical anomalies typically manifest as minor structural deviations rather than entirely new semantic categories, aggressively maximizing the distance between normal and abnormal features severely disrupts the continuous metric structure of the pre-trained latent space. This loss of feature continuity ultimately compromises the model's generalization to unseen open-world categories, highlighting the critical need for a continuity-preserving optimization strategy that enhances fine-grained discriminability without corrupting spatial metric integrity.
% 3 method
\section{Method}
\begin{figure*}[t] 
    \centering
    \includegraphics[width=\textwidth]{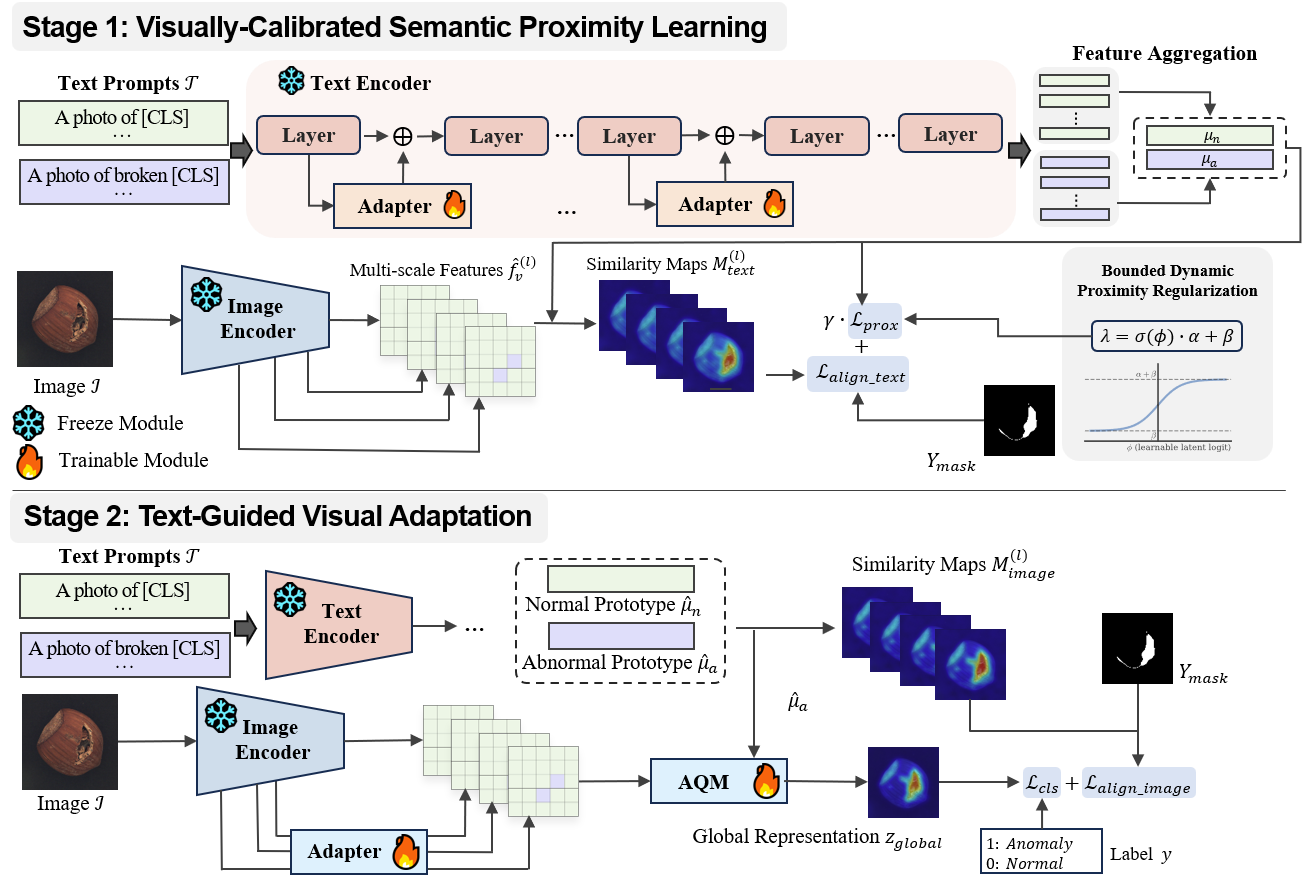}
    
    \caption{\textbf{Framework overview of Proximity-CLIP.} \textbf{Top, Stage 1:} Driven by actual visual variance, we dynamically learn a semantic margin $\lambda$ to calibrate the text prototypes $\mu_n$ and $\mu_a$, strictly preserving the pre-trained structural alignment. \textbf{Bottom, Stage 2:} Using the frozen anomalous prototype $\hat{\mu}_a$ as a semantic query, the Anomaly Query Module actively retrieves localized defect cues from contextual visual patches. The visual adapters are jointly optimized via classification and segmentation losses $\mathcal{L}_{cls}$ and $\mathcal{L}_{align\_image}$.}
    \label{fig:framework}
\end{figure*}

% 方法整体概述
As illustrated in Figure~\ref{fig:framework}, the proposed Proximity-CLIP is a two-stage parameter-efficient fine-tuning framework designed for Zero-Shot Anomaly Detection (ZSAD). In Stage 1, we introduce a visually-calibrated semantic proximity learning mechanism to solve the text-entanglement problem without disrupting the pre-trained metric space. In Stage 2, utilizing the calibrated anomalous prototype $\hat{\mu}_a$ as a driving semantic prior, we introduce an Anomaly Query Module (AQM) to actively aggregate localized defect cues from multi-scale visual patches.

% 3.1 
\subsection{Problem Setting and Geometric Dilemma}

\subsubsection{Task Formulation.} 
The goal of Zero-Shot Anomaly Detection (ZSAD) is to identify and localize anomalies in images without relying on any category-specific normal or abnormal training data. Formally, given an unseen test image $I \in \mathbb{R}^{H \times W \times 3}$, the model aims to predict an image-level anomaly score for global classification and a dense pixel-level anomaly map $M \in \mathbb{R}^{H \times W}$ for spatial localization, guided purely by textual semantic priors.

\subsubsection{Hyperspherical Semantic Prototypes.}
Following the standard paradigm of CLIP, the output image and text embeddings are intrinsically L2-normalized. Therefore, we naturally formulate the shared vision-language latent space as a $d$-dimensional unit hypersphere, denoted by:
\begin{equation}
\mathcal{H}^d = \{z \in \mathbb{R}^d \mid \left\|z\right\|_2 = 1\}
\end{equation}
In this constrained space, cross-modal alignment is exclusively governed by the cosine similarity $S(v, t) = v^\top t$. Building upon this hyperspherical metric formulation, the pre-trained embedding space establishes a robust structural alignment between visual and textual manifolds. For a specific object category, we define a set of normal textual prompts $\mathcal{T}_n$ and a set of anomalous prompts $\mathcal{T}_a$. To establish stable semantic references, we compute their L2-normalized semantic centers, referred to as the Normal Prototype $\mu_n$ and the Anomalous Prototype $\mu_a$:
\begin{equation}
\mu_n = \frac{\sum_{t \in \mathcal{T}_n} t}{\left\|\sum_{t \in \mathcal{T}_n} t\right\|_2}, \quad \mu_a = \frac{\sum_{t \in \mathcal{T}_a} t}{\left\|\sum_{t \in \mathcal{T}_a} t\right\|_2} \in \mathcal{H}^d
\end{equation}
The semantic proximity between the normal and anomalous states is strictly determined by their inner product $\mu_n^\top \mu_a$, which geometrically corresponds to the angle $\theta$ between them.

\subsubsection{The Geometric Dilemma.}
Despite their remarkable open-vocabulary capabilities, Vision-Language Models exhibit a strong object-centric bias, heavily prioritizing the overarching semantics of foreground categories while neglecting fine-grained local state descriptions. In our empirical observations, this bias dictates that the normal prototype $\mu_n$ and anomalous prototype $\mu_a$ directly extracted from CLIP maintain a substantially high cosine similarity. 

To decouple these entangled semantics, recent adapter-based methods attempt to strictly enforce orthogonality, mathematically driving the inner product $\mu_n^\top \mu_a$ toward zero. However, imposing such massive semantic distances introduces a severe structural dilemma. In physical reality, an anomalous instance is fundamentally a normal object perturbed by a localized defect, meaning their respective visual representations $v_n$ and $v_a$ remain inherently proximate. Forcing highly contiguous visual inputs to map onto drastically orthogonal semantic prototypes mathematically violates the geometric continuity of the pre-trained metric space, compromising the model's generalization on unseen categories. 

Therefore, their semantic distance should not be driven to an extreme zero. Instead, there exists a bounded, reasonable margin. This critical geometric dilemma fundamentally motivates the design of our proposed framework.

% 3.2
\subsection{Visually-Calibrated Semantic Proximity}

\subsubsection{Visually-Calibrated Semantic Margin.}
To resolve the aforementioned geometric inconsistency, we propose a visually-calibrated metric learning framework. In real-world scenarios, an anomalous instance is fundamentally a normal object perturbed by a localized defect. Consequently, their representations should not be separated into strictly orthogonal subspaces; instead, they should lie within a bounded topological neighborhood. To formalize this, we postulate the existence of a learnable semantic margin that characterizes an appropriate separation between normal and anomalous states. Rather than manually imposing a rigid heuristic distance, we utilize real-world image distributions to dynamically approximate this theoretical margin during the cross-modal alignment process.

\subsubsection{Norm-Preserving Residual Adapters.} 
During this stage, we freeze the CLIP encoders and insert lightweight MLPs as text adapters. To prevent the adapted features from disrupting the original magnitude distribution of the latent space, we introduce a norm-preserving residual blending mechanism. Given the hidden state $h$, the adapted output $\tilde{h}$ is constrained by:
\begin{equation}
\tilde{h} = \omega \cdot \left( \mathcal{A}(h) \frac{\left\|h\right\|_2}{\left\|\mathcal{A}(h)\right\|_2} \right) + (1 - \omega) \cdot h
\end{equation}
where $\mathcal{A}(\cdot)$ denotes the adapter projection and $\omega$ regulates the adaptation strength.

\subsubsection{Bounded Dynamic Proximity Regularization.} 
To explicitly instantiate the semantic margin, we introduce a learnable proximity target $\lambda$. Driven entirely by the true visual similarities between normal and abnormal distributions, this parameter dynamically learns an appropriate semantic margin. To ensure numerical stability, we design a bounded non-linear mapping using a learnable latent logit $\phi$ initialized to zero:
\begin{equation}
\lambda = \sigma(\phi) \cdot \alpha + \beta
\end{equation}
where $\sigma(\cdot)$ denotes the Sigmoid function, and the scaling scalars $\alpha$ and $\beta$ are defined to constrain the target similarity within a valid hyperspherical range. We continually force the actual cosine similarity between the current batch's text prototypes to approach this learned margin via a Mean Squared Error objective:
\begin{equation}
\mathcal{L}_{prox} = \frac{1}{B} \sum_{i=1}^{B} \left( \mu_{n, i}^\top \, \mu_{a, i} - \lambda \right)^2
\end{equation}

\subsubsection{Dual-Objective Joint Optimization.}
During text adaptation, the visual encoder is strictly frozen. The multi-scale patch features $\hat{f}_v^{(l)}$ are extracted from a predefined set of hierarchical layers $\mathcal{L}$ within the frozen visual branch. These features are densely aligned with the trainable text prototypes. The alignment loss $\mathcal{L}_{align\_text}$ is computed by applying a segmentation objective $\mathcal{L}_{seg}$ (comprising Focal and Dice losses) against the ground-truth mask $Y_{mask}$:
\begin{equation}
\mathcal{L}_{align\_text} = \sum_{l \in \mathcal{L}} \mathcal{L}_{seg}\left(\hat{f}_v^{(l)} [\mu_n, \mu_a]^\top, Y_{mask}\right)
\end{equation}
The overall optimization objective is $\mathcal{L}_{text} = \mathcal{L}_{align\_text} + \gamma \mathcal{L}_{prox}$, where $\gamma$ is a weighting hyperparameter. Through this joint optimization, gradients from the visual alignment explicitly update the text prototypes, while $\mathcal{L}_{prox}$ structurally couples them with $\lambda$. Consequently, the semantic margin is autonomously calibrated by the actual visual variance.

% 3.3
\subsection{Text-Guided Visual Adaptation}
Upon completing the text adaptation stage, the semantic text prototypes $(\hat{\mu}_n, \hat{\mu}_a)$ are strictly calibrated. During the subsequent visual adaptation stage, we freeze the entire text branch and leverage these state-aware semantic priors to actively guide the visual representations.

\subsubsection{Semantic-Guided Anomaly Querying.}
Standard visual adaptation typically processes image patches uniformly, which often causes subtle localized defects to be overshadowed by dominant normal backgrounds during global pooling. To bridge this gap, we exploit the calibrated anomalous prototype $\hat{\mu}_a$ as an explicit semantic prior to actively query the multi-scale visual patches. 

Specifically, we design a lightweight Anomaly Query Module (AQM) deployed across the hierarchical visual layers $\mathcal{L}$. For each layer $l \in \mathcal{L}$, $\hat{\mu}_a$ acts as the driving semantic query, while the explicitly adapted visual patch features $f_v^{(l)}$ serve as the contextual keys and values:
\begin{equation}
Q = \hat{\mu}_a, \quad K^{(l)} = f_v^{(l)}, \quad V^{(l)} = f_v^{(l)}
\end{equation}
where $f_{v}^{(l)}$ are the projected and L2-normalized visual tokens updated by the visual adapters. To mathematically realize this active retrieval, the AQM formulates a cross-modal interaction function, denoted by $\mathcal{F}_{attn}$. Rather than standard spatial aggregation, it inherently models the dense structural correlations between the explicit semantic prior and the local visual context. A residual connection is further integrated to stabilize the gradient flow:
\begin{equation}
z_{query}^{(l)} = \mathcal{F}_{attn}\left(Q, K^{(l)}, V^{(l)}\right) + \hat{\mu}_a
\end{equation}
To fuse the anomaly-aware features across all hierarchical scales, we perform a cross-layer mean pooling followed by L2-normalization to formulate the global anomaly representation $z_{global}$:
\begin{equation}
z_{global} = \frac{\bar{z}}{\left\|\bar{z}\right\|_2}, \quad \text{where} \quad \bar{z} = \frac{1}{|\mathcal{L}|} \sum_{l \in \mathcal{L}} z_{query}^{(l)}
\end{equation}

\subsubsection{Global-Local Joint Optimization.}
To achieve both fine-grained anomaly segmentation and image-level classification, the visual branch is optimized via a joint objective. For spatial grounding, the adapted multi-scale features are matched against the frozen text prototypes to calculate the segmentation alignment loss:
\begin{equation}
\mathcal{L}_{align\_image} = \sum_{l \in \mathcal{L}} \mathcal{L}_{seg}\left(f_v^{(l)} [\hat{\mu}_n, \hat{\mu}_a]^\top, Y_{mask}\right)
\end{equation}

For global anomaly classification, the pooled anomaly token $z_{global}$ is evaluated against the calibrated text prototypes. It is supervised by the image-level ground truth label $y \in \{0, 1\}$ using a cross-entropy loss $\mathcal{L}_{cls}$:
\begin{equation}
\mathcal{L}_{cls} = \text{CE}\left( \tau \cdot z_{global}^\top [\hat{\mu}_n, \hat{\mu}_a], \, y \right)
\end{equation}
where $\tau$ is the learnable temperature scaling parameter. Ultimately, the overall loss for the visual adaptation stage is formulated as $\mathcal{L}_{image} = \mathcal{L}_{align\_image} + \mathcal{L}_{cls}$. This dual-level optimization ensures that the visual adapters effectively uncover localized defects while forming a robust categorical representation.

% 4 experiment
\section{Experiments}
\label{sec:experiments}

%----------------------------实验设置-------------------------------------%
\subsection{Experimental Setup}

\textbf{Datasets.} The proposed framework is evaluated on seven diverse datasets originating from both the industrial and medical domains. For industrial anomaly detection, four standard benchmarks are utilized: MVTec-AD~\cite{bergmann2019mvtec}, VisA~\cite{zou2022spot}, BTAD~\cite{mishra2021vt}, and MPDD~\cite{jezek2021deep}. Within the medical domain, three additional datasets are incorporated: Brain MRI, Liver CT, and Retina OCT from BMAD~\cite{bao2024bmad}. Adhering to the cross-dataset zero-shot evaluation protocol established in prior studies~\cite{ma2025aa}, the VisA dataset is employed as the training source for the evaluation of MVTec-AD, BTAD, MPDD, as well as the three medical datasets. Conversely, for the evaluation conducted on the VisA dataset, the model is trained exclusively on MVTec-AD.

\noindent\textbf{Evaluation Metrics.} Following prior works~\cite{ma2025aa, zhou2023anomalyclip,huang2024adapting,jeong2023winclip}, we use the Area Under the Receiver Operating Characteristic Curve (AUROC) at both the image and pixel levels as our primary evaluation metrics. The Image AUROC measures the image-level anomaly classification performance, while the Pixel AUROC evaluates the precision of spatial anomaly localization by comparing pixel-wise predictions against ground-truth masks.

\noindent\textbf{Implementation Details.} Proximity-CLIP is implemented based on the OpenCLIP architecture, utilizing the ViT-L/14 visual backbone. All input images are uniformly resized to a resolution of $518 \times 518$. During the training phase, the Adam optimizer is employed to update the parameters of the network. The text and image adaptation branches are optimized with learning rates of $1\times10^{-5}$ and $5\times10^{-4}$, respectively. Furthermore, a dedicated learning rate of 0.01 is assigned to the learnable similarity module. To ensure numerical stability and prevent extreme geometric deviations, the scaling scalars $\alpha$ and $\beta$ within the bounded dynamic proximity regularization are empirically set to 0.65 and 0.3, respectively. This configuration strictly constrains the dynamic margin $\lambda$ to a valid mathematical interval of $[0.3, 0.95]$. The optimization process spans 10 epochs for the first adaptation stage and 25 epochs for the second stage. All training procedures are conducted on a single NVIDIA RTX A6000 GPU, while the testing phase is executed on an NVIDIA RTX 4090 GPU.

%---------------------------主要结果----------------------------------------%
\subsection{Main Results}

% ================= pixel-level的AUROC表 =================
\begin{table*}[!t]
\centering
\caption{Quantitative comparison of anomaly localization performance (Pixel AUROC $\uparrow$) across four industrial and three medical datasets. The best and second-best results are highlighted in \textbf{bold} and \underline{underlined}, respectively.}
\setlength{\tabcolsep}{4pt} 
\resizebox{\textwidth}{!}{
\begin{tabular}{c | l | c c c c c c c | c}
\toprule
Domain & Dataset & CLIP & WinCLIP & VAND & MVFA-AD & AnomalyCLIP & AdaCLIP & AA-CLIP & \makebox[1.5cm][c]{Ours} \\
\midrule
\multirow{4}{*}{Industrial} 
 & BTAD & 30.6 & 32.8 & 91.1 & 90.1 & 93.3 & 90.8 & \underline{97.0} & \textbf{97.0} \\ 
 & MPDD & 62.1 & 95.2 & 94.9 & 94.5 & 96.2 & \textbf{96.6} & \underline{96.5} & 96.2 \\
 & MVTec-AD & 38.4 & 85.1 & 87.6 & 84.9 & 91.1 & 89.9 & \underline{91.8} & \textbf{92.0} \\ 
 & VisA & 46.6 & 79.6 & 94.2 & 93.4 & 95.4 & \underline{95.5} & 94.7 & \textbf{95.7} \\
\midrule
\multirow{3}{*}{Medical} 
 & Brain MRI & 68.3 & 86.0 & 94.5 & 95.6 & \textbf{96.2} & 93.9 & 95.3 & \textbf{96.2} \\
 & Liver CT & 90.5 & 96.2 & 95.6 & \underline{96.8} & 93.9 & 94.5 & \textbf{97.6} & \underline{96.8} \\
 & Retina OCT & 21.3 & 80.6 & 88.5 & 90.9 & 92.6 & 88.5 & \underline{95.5} & \textbf{95.9} \\
\midrule
\multicolumn{2}{c|}{Average} & 51.0 & 79.4 & 92.3 & 92.3 & 94.1 & 92.8 & \underline{95.5} & \textbf{95.7} \\
\bottomrule
\end{tabular}
}
\label{tab:pixel_auroc}
\end{table*} % pixel-level结果表格
% ================= image-level的AUROC表=================
\begin{table*}[!t]
\centering
\caption{Quantitative comparison of image-level anomaly classification performance (Image AUROC $\uparrow$). The best and second-best results are highlighted in \textbf{bold} and \underline{underlined}, respectively.}
\setlength{\tabcolsep}{4pt} % 缩小列间距，让左边紧凑
\resizebox{\textwidth}{!}{
\begin{tabular}{c | l | c c c c c c | c}
\toprule
Domain & Dataset & CLIP & WinCLIP & MVFA-AD & AnomalyCLIP & AdaCLIP & AA-CLIP & \makebox[1.5cm][c]{Ours} \\
\midrule
\multirow{4}{*}{Industrial} 
 & BTAD & 73.6 & 68.2 & \underline{94.3} & 85.3 & 90.9 & \textbf{94.8} & 92.3 \\ 
 & MPDD & 73.0 & 63.6 & 70.9 & 73.7 & 72.1 & \underline{73.8} & \textbf{74.1} \\
 & MVTec-AD & 86.1 & \underline{91.8} & 86.6 & 90.9 & 90.0 & 90.0 & \textbf{92.2} \\ 
 & VisA & 66.4 & 78.0 & 76.5 & 82.1 & \textbf{84.3} & 78.3 & \underline{82.9} \\
\midrule
\multirow{3}{*}{Medical} 
 & Brain MRI & 58.8 & 66.5 & 70.9 & \textbf{83.3} & \underline{80.2} & 77.6 & 77.2 \\
 & Liver CT & 54.7 & \underline{64.2} & 63.0 & 61.6 & \underline{64.2} & \textbf{66.8} & 64.0 \\
 & Retina OCT & 65.6 & 42.5 & 77.3 & 75.7 & \underline{82.7} & \underline{82.7} & \textbf{84.3} \\
\midrule
\multicolumn{2}{c|}{Average} & 68.3 & 67.8 & 77.1 & 78.9 & \underline{80.6} & \underline{80.6} & \textbf{81.0} \\
\bottomrule
\end{tabular}
}
\label{tab:image_auroc}
\end{table*} % image-level结果表格

To rigorously evaluate the effectiveness of Proximity-CLIP, we compare it against zero-shot anomaly detection baselines, including CLIP~\cite{radford2021learning}, WinCLIP~\cite{jeong2023winclip}, VAND~\cite{chen2023zero}, MVFA-AD~\cite{huang2024adapting}, AnomalyCLIP~\cite{zhou2023anomalyclip}, AdaCLIP~\cite{cao2024adaclip}, and AA-CLIP~\cite{ma2025aa}. The comparison is conducted systematically from two perspectives: fine-grained anomaly localization and holistic anomaly classification.

\subsubsection{Anomaly Localization.}
Table~\ref{tab:pixel_auroc} presents the quantitative comparison of pixel-level anomaly localization performance. Proximity-CLIP effectively delineates anomalous regions and achieves highly competitive results compared to existing methods. Specifically, across all seven evaluated datasets from both the industrial and medical domains, our method achieves an exceptional average Pixel AUROC of 95.7\%. Notably, it maintains robust localization accuracy even on complex medical datasets like Brain and Retina, which feature highly subtle defect patterns. This consistent precision demonstrates that our proposed visually-calibrated semantics and active querying mechanism work synergistically to effectively mitigate the dilution of subtle anomalies during spatial aggregation.

\subsubsection{Anomaly Classification.} 
Table~\ref{tab:image_auroc} summarizes the image-level anomaly classification results. Consistent with the localization performance, Proximity-CLIP exhibits strong classification capabilities, attaining the highest overall average classification score of $81.0\%$ and outperforming all baseline models. The evaluation results across diverse semantic domains validate the robust cross-domain generalization capability of the proposed framework. This robustness confirms that our continuity-preserving paradigm successfully prevents global categorical representations from being distorted, ensuring reliable image-level anomaly recognition.

%---------------------------深入分析----------------------------------------%
\subsection{In-Depth Analysis}

% 语义临近校准(混淆矩阵分析)
\subsubsection{Semantic Proximity Calibration.}
% 混淆矩阵图
\begin{figure*}[!t] % figure* 用于双栏排版中的跨栏大图
    \centering
    \includegraphics[width=0.95\textwidth]{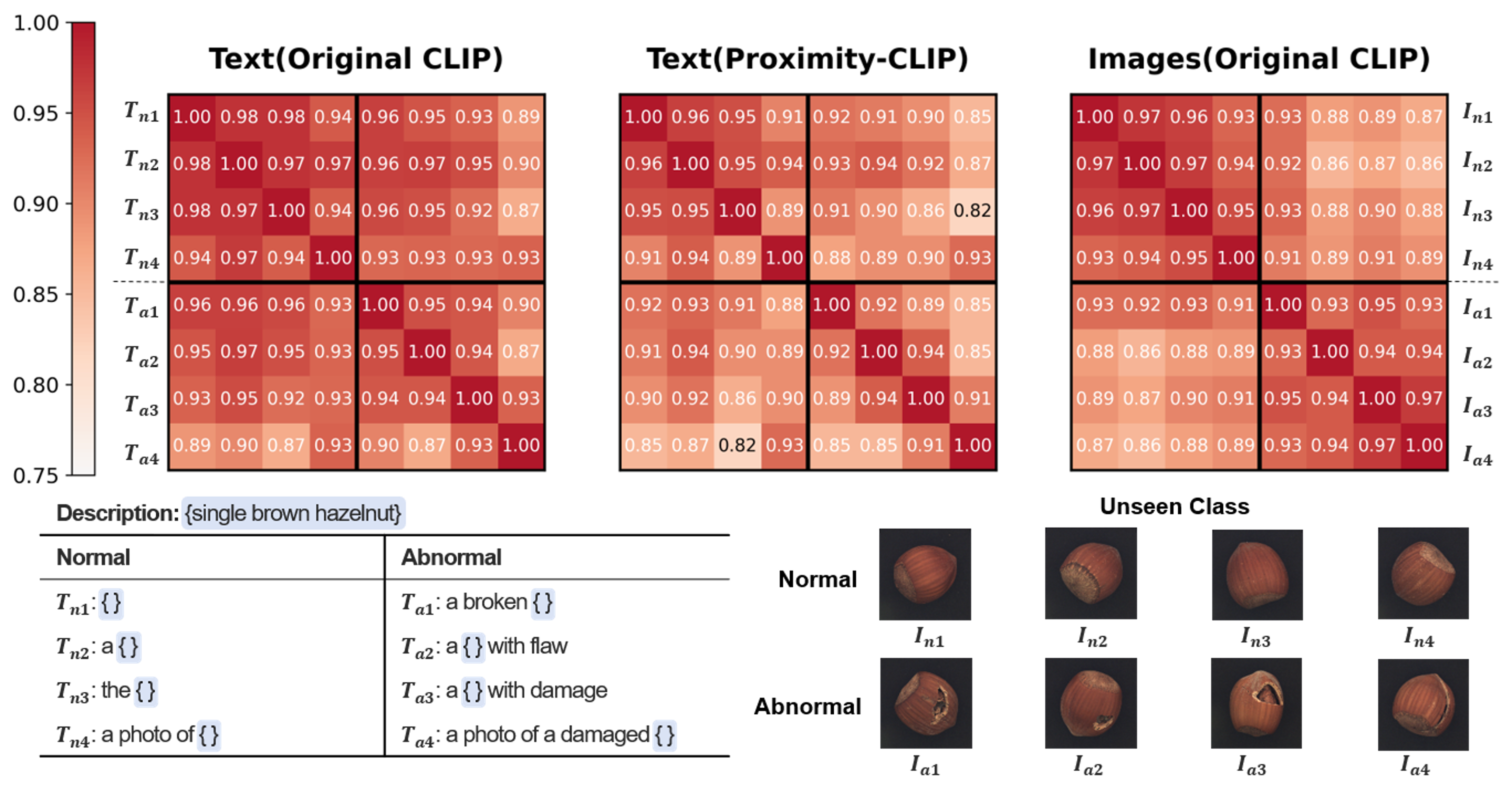} % 略小于全文宽度，看起来更美观
    \caption{
        Cosine similarity matrices illustrating semantic proximity calibration on unseen classes. The leftmost matrix reveals severe semantic entanglement in original CLIP text embeddings, whereas the rightmost matrix exhibits the intrinsic similarity distribution of actual images. The middle matrix demonstrates that our Stage 1 calibration effectively reshapes textual boundaries to mirror the authentic visual distribution, maintaining appropriate semantic proximity instead of forcing heuristic orthogonal separation.
    }
    \label{fig:confuse_matrix}
\end{figure*}
Figure~\ref{fig:confuse_matrix} visualizes the cosine similarity matrices of normal and anomalous embeddings from unseen classes to validate our Stage 1 calibration. The leftmost matrix reveals severe semantic entanglement in the original CLIP text embeddings, where excessively high similarities between normal and anomalous texts hinder zero-shot discrimination. Conversely, the rightmost matrix reveals the intrinsic similarity distribution of the actual visual features. Remarkably, the middle matrix demonstrates that our calibrated text-to-text similarity closely mirrors this authentic visual distribution. Rather than forcing complete orthogonal separation that disrupts the pre-trained metric space, our approach reshapes text boundaries to maintain an appropriate, visually-aligned distance. This structural calibration resolves semantic entanglement and preserves metric integrity, establishing an accurate prior for subsequent active visual retrieval.

% t-sne图和可视化图
% \begin{figure}[!t]
%     \centering
%     % ================= 左侧：t-SNE 图 =================
%     % 将 [b] 改为 [c]，垂直居中对齐
%     \begin{minipage}[c]{0.48\textwidth} 
%         \centering
%         \includegraphics[width=\linewidth]{images/t_sne.png}
%         \caption{ t-SNE visualizations of feature embeddings across seen and unseen classes.}
%         \label{fig:tsne}
%     \end{minipage}
%     \hfill
%     % ================= 右侧：消融实验表 =================
%     % 同样将 [b] 改为 [c]
%     \begin{minipage}[c]{0.48\textwidth}
%         \centering
%         \makeatletter\def\@captype{table}\makeatother
%         \renewcommand{\arraystretch}{0.95}
%         \caption{ Ablation study of Semantic Calibration (SC) and the Anomaly Query Module (AQM).}
%         \label{tab:ablation}
        
%         \scriptsize 
%         \setlength{\tabcolsep}{4pt} 
%         \begin{tabular}{@{} l c c c c @{}}
%         \toprule
%         \multirow{2}{*}{Variant} & \multicolumn{2}{c}{Industrial} & \multicolumn{2}{c}{Medical} \\
%         \cmidrule(lr){2-3} \cmidrule(lr){4-5}
%         & Pixel & Image & Pixel & Image \\
%         \midrule
%         Baseline &  57.5 & 60.3 & 55.5 & 44.3 \\
%         w/o SC  & 91.9 & 70.4 & 92.9 & 73.9 \\
%         w/o AQM   & 68.8 & 69.1 & 84.4 & 53.5 \\
%         Ours     & \textbf{95.1} & \textbf{86.2} & \textbf{96.3} & \textbf{75.2} \\
%         \bottomrule
%         \end{tabular}
%     \end{minipage}
% \end{figure}

% 图表环境排版
\begin{figure}[!t]
    \centering
    % ================= 左侧：t-SNE 图 =================
    \begin{minipage}[c]{0.48\textwidth} 
        \centering
        \includegraphics[width=\linewidth]{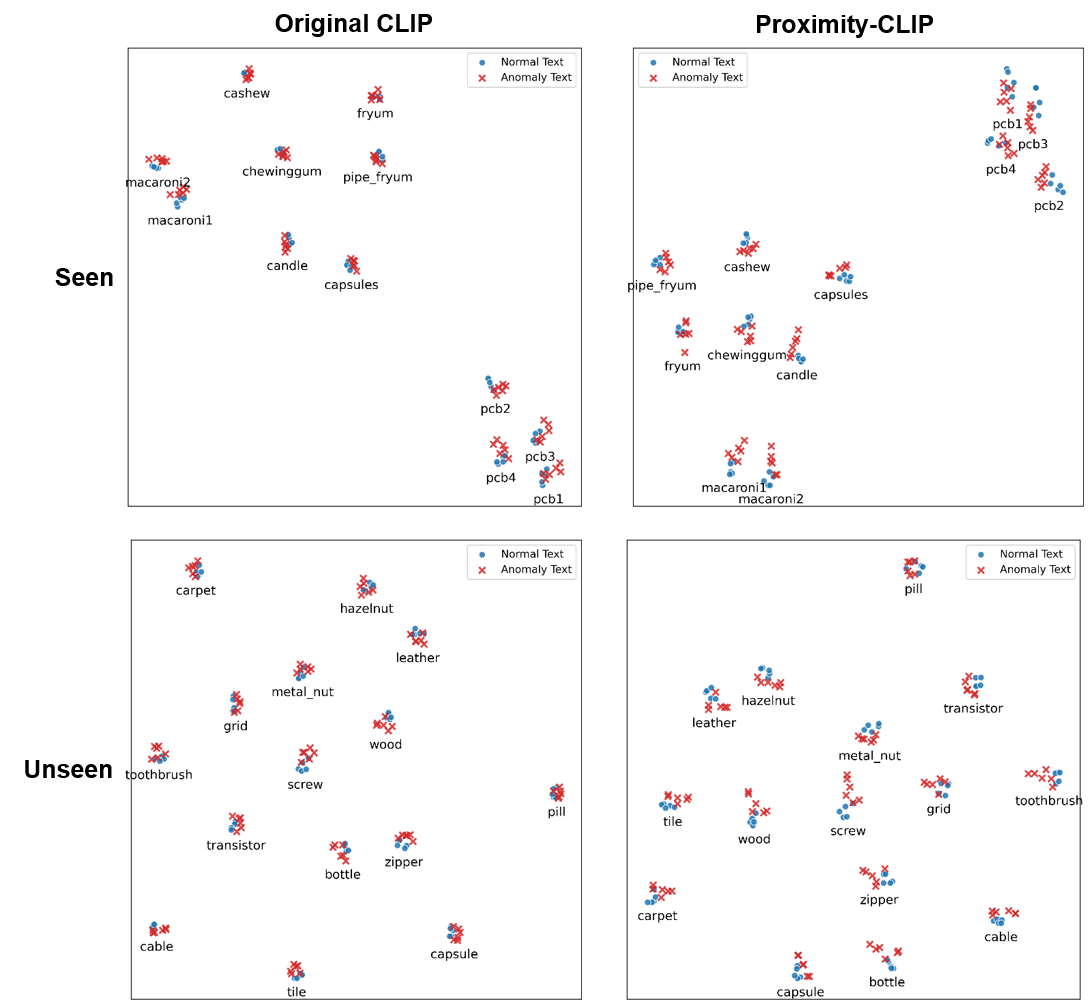}
        \caption{ t-SNE visualizations of feature embeddings across seen and unseen classes.}
        \label{fig:tsne}
    \end{minipage}
    \hfill
    % ================= 右侧：整合后的消融实验表 =================
    \begin{minipage}[c]{0.46\textwidth} % 1. 将右侧容器宽度从 0.48 稍微缩小到 0.46
        \centering
        \makeatletter\def\@captype{table}\makeatother
        \renewcommand{\arraystretch}{0.95}
        \caption{ Comprehensive ablation study of our proposed modules, Norm-Preserving (NP) mechanism, and retrieval strategies.}
        \label{tab:ablation}
        
        \tiny % 2. 将 \scriptsize 改小一号为 \tiny
        \setlength{\tabcolsep}{2pt} % 3. 列宽从 3pt 进一步收缩为 2pt，防止内容过长撑出边界
        \begin{tabular}{@{} l c c c c @{}}
        \toprule
        \multirow{2}{*}{Variant} & \multicolumn{2}{c}{Industrial (\%)} & \multicolumn{2}{c}{Medical (\%)} \\
        \cmidrule(lr){2-3} \cmidrule(lr){4-5}
        & Pixel & Image & Pixel & Image \\
        \midrule
        Baseline & 57.5 & 60.3 & 55.5 & 44.3 \\
        w/o SC  & 91.9 & 70.4 & 92.9 & 73.9 \\
        w/o AQM   & 68.8 & 69.1 & 84.4 & 53.5 \\
        \midrule
        w/o NP (Text) & 95.0 & 82.7 & 95.5 & 72.3 \\
        w/o NP (Image) & 94.6 & 81.9 & 95.9 & 68.5 \\
        w/o NP (Both) & 94.5 & 82.0 & 95.8 & 70.8 \\
        \midrule
        Multi-scale (w/o AQM) & 93.8 & 85.4 & 95.5 & 73.0 \\
        \midrule
        Ours (Full Model) & \textbf{95.1} & \textbf{86.2} & \textbf{96.3} & \textbf{75.2} \\
        \bottomrule
        \end{tabular}
    \end{minipage}
\end{figure} % tsne图+消融实验表
\begin{figure}[!t] % 如果需要跨越左右双栏，改为 \begin{figure*}[!t]
    \centering
    \includegraphics[width=\linewidth]{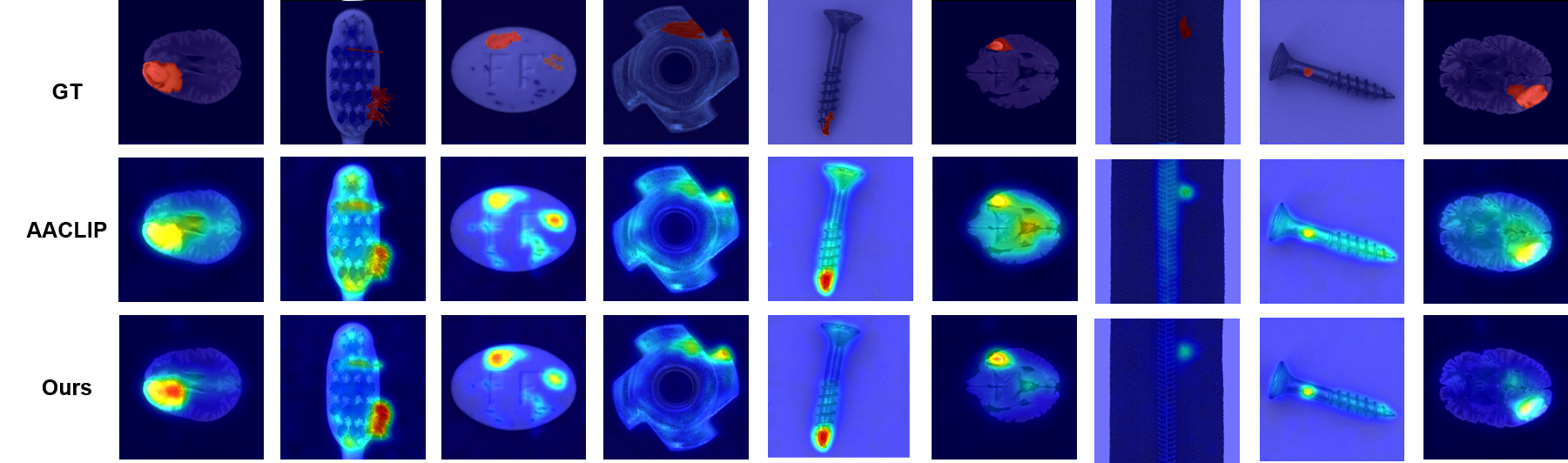}
    \caption{Qualitative comparison of anomaly localization maps between the previous state-of-the-art method AA-CLIP and our Proximity-CLIP. Proximity-CLIP exhibits clear superiority in accurately localizing anomalies.}
    \label{fig:vis}
\end{figure} % 可视化图
% t-sne
\subsubsection{Preserving Metric Space Integrity.}
Figure~\ref{fig:tsne} provides t-SNE visualizations of the feature embeddings for both seen and unseen classes. In the original CLIP latent space, normal and anomalous samples within the same category are heavily entangled. In contrast, Proximity-CLIP achieves distinct intra-class separation between normal and anomalous features. This separation is accomplished while preserving the pretrained inter-class topological structure, as different categories remain clearly demarcated. Geometrically, our method distinguishes normal and anomalous semantics through a controlled angular shift in cosine similarity rather than forcing them into orthogonal or uncorrelated spaces. This continuity-preserving mechanism prevents distortion of the original hyperspherical metric space, maintaining zero-shot generalization capabilities on unseen categories.

% 可视化
\subsubsection{Qualitative Anomaly Localization.}
Figure~\ref{fig:vis} presents a qualitative comparison of anomaly localization heatmaps across diverse datasets. Compared with the previous state-of-the-art method AA-CLIP, Proximity-CLIP demonstrates superior capability in delineating anomalous regions, particularly for subtle and fine-grained defects. As illustrated in the visualizations, our method precisely pinpoints the anomalous areas while effectively mitigating background noise interference. By leveraging the visually-calibrated anomalous semantics as an active query, our cross-attention adapter acts as a precise semantic spotlight. This mechanism robustly suppresses normal background representations and explicitly highlights complex structural anomalies. These visual results consistently corroborate our quantitative evaluations and validate the superiority of our active anomaly retrieval paradigm.

%---------------------------消融实验----------------------------------------%
\subsection{Ablation Study}

To evaluate the individual contributions of our proposed modules, we conduct an ablation study as detailed in Table~\ref{tab:ablation}. The baseline variant relies on the original CLIP architecture with standard manual prompts and passive global matching, yielding sub-optimal performance across both domains. The \textbf{w/o AQM} variant introduces only the stage-one Semantic Calibration. While this calibration improves basic text-image alignment, raising the Medical pixel AUROC from 55.5\% to 84.4\%, it still suffers from passive feature dilution in dense prediction tasks. Conversely, the \textbf{w/o SC} variant employs the Anomaly Query Module but omits the stage-one calibration. Although active retrieval significantly boosts pixel-level localization, the absence of a visually-calibrated semantic margin leaves the normal and anomalous text prototypes highly entangled. This lack of decoupled semantic priors bottlenecks image-level classification, causing the Industrial image AUROC to drop substantially from 86.2\% to 70.4\%. 

Furthermore, we investigate the impact of our \textbf{Norm-Preserving (NP) mechanism}. Removing this mechanism (for Text, Image, or Both) causes the Image AUROC to plummet across both domains (e.g., a drop of 4.4\% in Medical and 4.2\% in Industrial when removing both). This effectively confirms its critical necessity in stabilizing pre-trained feature magnitudes during metric adaptation. 

Finally, to validate the \textbf{expressiveness of the AQM}, we compare it against a multi-scale pooling baseline. Rather than acting as a rigid defect template, our calibrated prototype serves as a generalized structural deviation prior. Actively querying this prior via cross-attention dynamically retrieves diverse, localized defects, outperforming passive multi-scale pooling by boosting Industrial Pixel (+1.3\%) and Medical Image AUROC (+2.2\%). Ultimately, our full Proximity-CLIP integrates all components to achieve the best overall performance among all evaluated variants. This validates that the visually-calibrated semantics and the active visual retrieval mechanism are highly synergistic and indispensable for robust zero-shot anomaly detection.

% 5 conclusion
\section{Conclusion}
\label{conclusion}

In this paper, we presented Proximity-CLIP, a framework that reformulates zero-shot anomaly detection through text-guided semantic proximity learning. We showed that strictly enforcing orthogonality between text prototypes creates a geometric conflict that disrupts pre-trained structural continuity. To address this, we developed a visually-calibrated semantic proximity learning mechanism. Using bounded dynamic regularization, this approach learns a semantic margin to ensure discriminative separation while preserving the original metric integrity. Based on this calibrated space, we designed an Anomaly Query Module (AQM). The AQM uses the anomalous prototype as a semantic query to retrieve localized defect cues from visual patches, preventing the dilution of subtle anomalies that occurs during standard global pooling. Experiments on industrial and medical benchmarks demonstrate that Proximity-CLIP achieves state-of-the-art zero-shot performance with minimal architectural changes. Our continuity-preserving approach and active retrieval strategy offer a practical foundation for future work in open-world dense prediction tasks.

% References should start immediately after the main text, but can continue past p.\ 14 if needed. 
% \clearpage  % TODO FINAL: This \clearpage needs to be removed from both review and camera-ready versions.
% Acknowledgements
\section*{Acknowledgements}
This work was supported in part by the National Key Research and Development Program of China under Grant 2022YFB3303800.

\bibliographystyle{splncs04}
\bibliography{main}

% 补充材料接到正文 PDF 末尾（arXiv / 合并阅读用）
% suppl.tex
% 不要单独编译。在 main.tex 参考文献之后 \input{suppl}

\clearpage

% 章节编号改为 A, B, C, ...
\renewcommand{\thesection}{\Alph{section}}
\renewcommand{\thesubsection}{\thesection.\arabic{subsection}}
\renewcommand{\thesubsubsection}{\thesubsection.\arabic{subsubsection}}

% 图表编号改为 A.1, B.1, ...
\renewcommand{\thefigure}{\thesection.\arabic{figure}}
\renewcommand{\thetable}{\thesection.\arabic{table}}
% \renewcommand{\thealgorithm}{\thesection.\arabic{algorithm}}

% 公式编号改为 A.1, B.1, ...（和正文合并成一个 PDF 时避免与正文 (1)(2) 撞号）
\renewcommand{\theequation}{\thesection.\arabic{equation}}

\makeatletter
\@addtoreset{figure}{section}
\@addtoreset{table}{section}
\@addtoreset{equation}{section}
% \@addtoreset{algorithm}{section}
\makeatother

\setcounter{section}{0}
\setcounter{figure}{0}
\setcounter{table}{0}
\setcounter{equation}{0}

\vspace*{1em}
\begin{center}
{\LARGE\bfseries Supplementary Material\par}
\end{center}
\vspace{1.2em}

\section{More Quantitative Results }

To further evaluate the generalization of our proposed framework, we provide additional experimental results on four challenging medical datasets. It is worth noting that these datasets consist exclusively of anomalous images without any normal samples. Consequently, image-level classification metrics cannot be formulated. We solely report the pixel-level anomaly localization performance (Pixel AUROC). As presented in Table~\ref{tab:medical_pixel_auroc}, Proximity-CLIP consistently outperforms previous methods across all evaluated datasets. These results demonstrate the effectiveness of our approach in accurately delineating abnormal regions, further validating its robust capability to focus on fine-grained anomalies within complex anatomical structures.

% ================= Medical datasets AUROC表 =================
\begin{table*}[!h]
\centering
\caption{Quantitative comparison of anomaly localization performance (Pixel AUROC $\uparrow$) across four medical datasets. The best and second-best results are highlighted in \textbf{bold} and \underline{underlined}, respectively.}
\setlength{\tabcolsep}{4pt} 
\resizebox{\textwidth}{!}{
\begin{tabular}{c | l | c c c c c c c | c}
\toprule
Domain & Dataset & CLIP & WinCLIP & VAND & MVFA-AD & AnomalyCLIP & AdaCLIP & AA-CLIP & \makebox[1.5cm][c]{Ours} \\
\midrule
\multirow{4}{*}{Medical} 
 & ColonDB & 49.5 & 51.2 & 78.2 & 78.4 & 82.9 & 80.0 & \underline{84.0}& \textbf{84.3} \\
 & ClinicDB & 47.5 & 70.3 & 85.1 & 83.9 & 85.0 & 85.9 & \underline{89.9} & \textbf{90.6} \\
 & Kvasir & 44.6 & 69.7 & 80.3 & 81.9 & 81.9 & 86.4 & \underline{87.2}& \textbf{88.1}  \\
 & CVC-300 & 49.9 & - & 92.8 & 82.6 & 95.4 & 92.9 & \underline{96.4} & \textbf{96.7} \\
\midrule
\multicolumn{2}{c|}{Average} & 47.9 & 63.7 & 84.1 & 81.7 & 86.3 & 86.3 & \underline{89.4} & \textbf{89.9} \\
\bottomrule
\end{tabular}
}
\label{tab:medical_pixel_auroc}
\end{table*}

\section{Extended Analysis}
This section provides further details regarding the architectural design and theoretical foundations of Proximity-CLIP.

\subsection{Design Rationale of Bounded Dynamic Proximity Regularization}This subsection presents additional justification for the formulation of Bounded Dynamic Proximity Regularization. Specifically, it addresses the parameterization of the dynamic margin $\lambda$ and the selection of the scaling scalars $\alpha$ and $\beta$.
\subsubsection{Rationale for Encapsulating the Learnable Margin $\lambda$}
Although instantiating the target margin $\lambda$ directly as a standard learnable scalar parameter might appear straightforward, this approach introduces significant risks to the stability of the training process. The target $\lambda$ is designed to guide the cosine similarity $\mu_{n, i}^\top \, \mu_{a, i}$ between normal and abnormal textual prototypes. By mathematical definition, cosine similarity is strictly bounded within the interval $[0, 1]$.If $\lambda$ is treated as an unconstrained parameter ($\lambda \in \mathbb{R}$), unhindered gradient updates could easily push the value outside this valid range (\eg, $\lambda > 1$ or $\lambda < 0$). Should this occur, the Mean Squared Error objective ($\mathcal{L}_{prox}$) forces the model to optimize toward a mathematically impossible target. Such a scenario leads to severe gradient instability.To prevent this, the learning process is encapsulated within a bounded non-linear mapping. By optimizing a latent logit $\phi$, which is initialized to zero, and applying the Sigmoid function $\sigma(\phi) \in (0, 1)$, the base optimization space is inherently restricted. The subsequent affine transformation via the scalars $\alpha$ and $\beta$ maps this stable $(0, 1)$ output to a precisely controlled and mathematically valid sub-interval. This mechanism ensures that the target $\lambda$ consistently remains geometrically meaningful throughout the optimization process.

\begin{figure}[!h] % [t] indicates placing the figure at the top of the page
    \centering
    \includegraphics[width=0.65\linewidth]{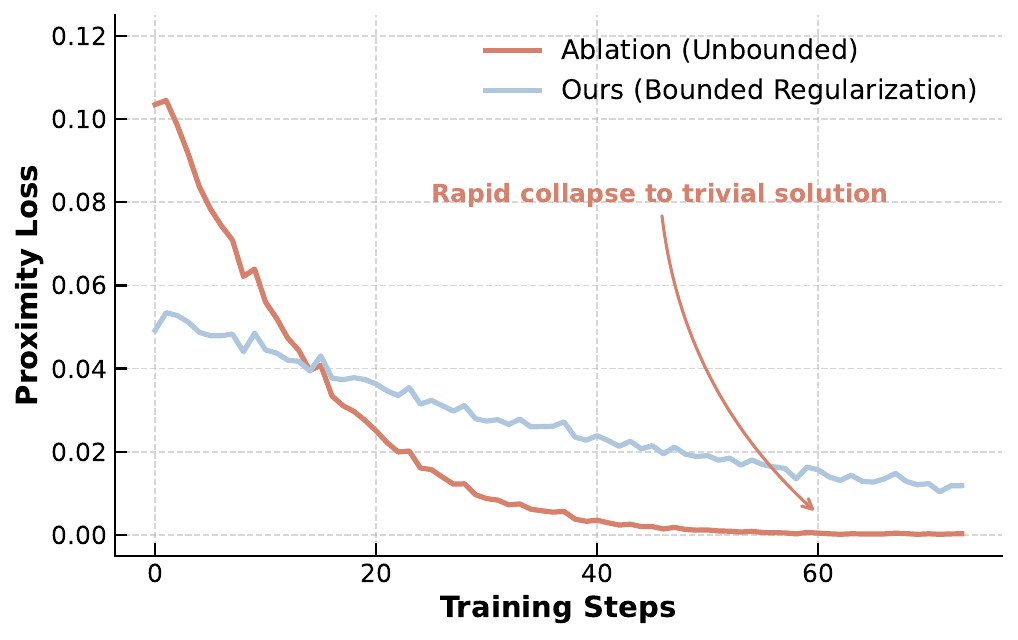}
    \caption{\textbf{Convergence analysis of the proximity loss $\mathcal{L}_{prox}$.} The comparison between the unbounded scalar $\lambda$ and our Bounded Regularization demonstrates that the proposed encapsulation effectively prevents optimization shortcuts. While the unbounded $\mathcal{L}_{prox}$ collapses instantly to a trivial solution, the bounded variant maintains a steady and non-zero convergence, ensuring the learning of discriminative textual representations.}
    \label{fig:loss_convergence}
\end{figure}

\subsubsection{Preventing Optimization Shortcuts via Bounded Encapsulation.} To demonstrate the necessity of the proposed bounded non-linear mapping for $\lambda$, we tracked the convergence of the proximity loss ($\mathcal{L}_{prox}$) during the initial text adaptation phase, as illustrated in Figure \ref{fig:loss_convergence}. When $\lambda$ is treated as an unconstrained scalar, $\mathcal{L}_{prox}$ collapses toward zero almost instantly. Rather than indicating successful learning, this rapid decline represents a classic manifestation of an optimization shortcut, \ie, a trivial solution. Since the unbounded $\lambda$ possesses total degrees of freedom, the network minimizes the MSE loss by aggressively adjusting $\lambda$ to track the dot product $\mu_{n}^\top \, \mu_{a}$, which inevitably causes the textual prototypes to merge and leads to feature collapse.

In contrast, the Bounded Regularization mechanism mathematically encapsulates $\lambda$ within a Sigmoid function and defined scalars. This structural design functions as a critical gradient damper, providing necessary optimization resistance that prevents the model from exploiting trivial solutions. The steady, non-zero convergence of the bounded $\mathcal{L}_{prox}$ demonstrates that the network effectively learns distinct and discriminative textual representations, thereby strictly preserving the geometric margin required for downstream anomaly detection.

\subsection{Sensitivity Analysis of Margin Bounds}
\begin{figure}[htb]
  \centering
  \includegraphics[width=0.98\linewidth]{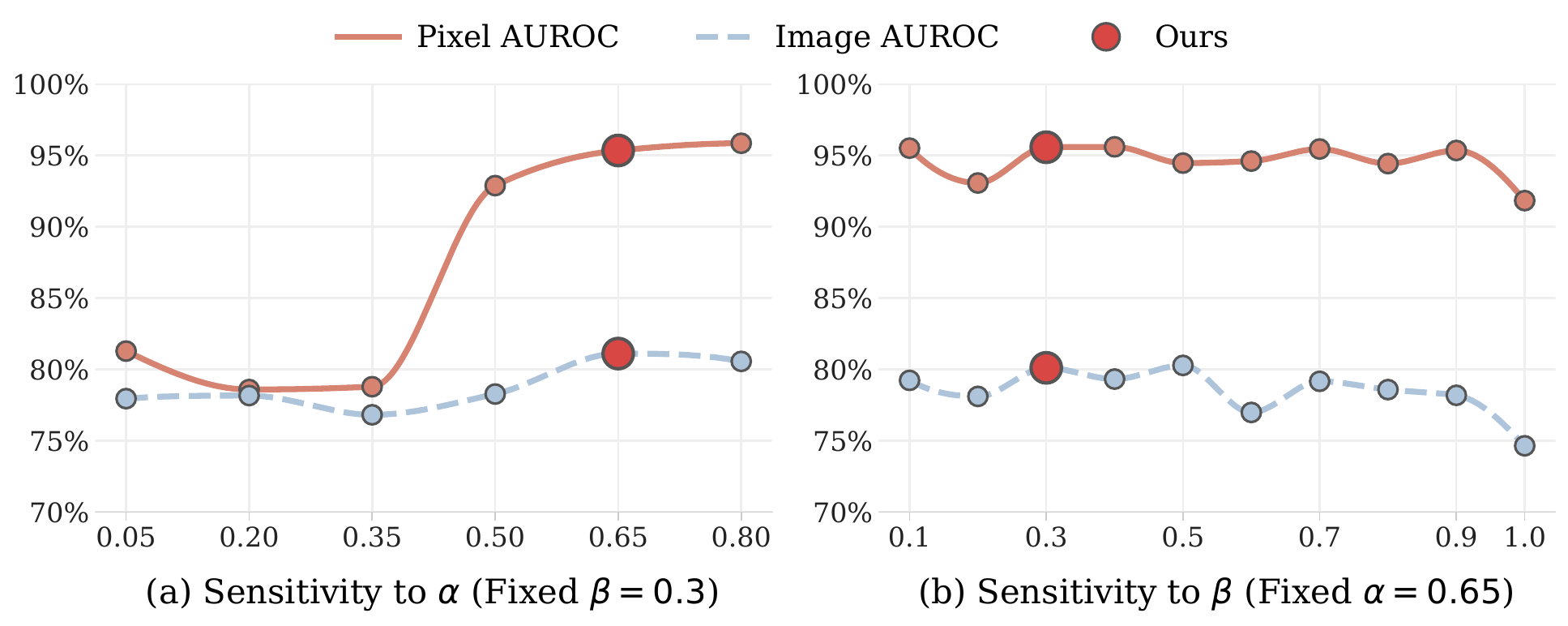}
  \caption{Sensitivity analysis of the hyperparameter bounds $\alpha$ and $\beta$, averaged over all six evaluated datasets.}
  \label{fig:sensitivity}
\end{figure}

To further validate the robustness of the dynamic margin bounds ($\alpha$ and $\beta$), we conduct a comprehensive sensitivity analysis. The theoretical necessity of encapsulating these parameters is elaborated in the preceding subsection. Figure~\ref{fig:sensitivity} illustrates the performance variations across different configurations of $\alpha$ and $\beta$.

\noindent\textbf{Validation of the Geometric Dilemma ($\alpha$):} As shown in Figure~\ref{fig:sensitivity} (left), with the lower bound $\beta$ fixed at $0.3$, the detection performance remains highly stable for $\alpha \ge 0.5$. However, the performance degrades sharply when $\alpha \le 0.35$, corresponding to cases where the upper bound is restricted to 0.65 or lower. This empirical observation explicitly corroborates the \textbf{Geometric Dilemma} identified in the main manuscript: enforcing strict orthogonality severely disrupts the pre-trained structural continuity of the vision-language representations.

\noindent\textbf{Robustness to Lower Bound ($\beta$):} Fixing $\alpha=0.65$ (Fig.~\ref{fig:sensitivity}, right), the model exhibits exceptional stability. As long as the upper bound ($\alpha$) is wide enough to cover true visual similarities, the network dynamically adapts to the optimal margin. Performance only declines explicitly when the lower bound $\beta$ approaches 1.0, as this violates intrinsic physical similarities and restricts dynamic expansion. Thus, our setting (0.65, 0.3) simply constructs a safe, physically grounded constraint requiring no domain-specific tuning.

\subsection{Computational Efficiency and Overhead Analysis}
In this section, we provide a detailed analysis of the computational overhead to demonstrate the high parameter efficiency of the proposed framework, as summarized in \textbf{Table~\ref{tab:efficiency}}. Specifically, the proposed method introduces only $\sim$22.0M trainable parameters, which constitutes merely 5\% of the total parameters of the backbone model. Furthermore, the \textit{per-epoch} training times of the proposed approach are highly competitive with those of AA-CLIP: Stage 1 is slightly faster, whereas Stage 2 introduces a minimal computational overhead of only 4\%. 

% \begin{table}[h]
%   \centering
%   \caption{Computational efficiency comparison. Proximity-CLIP achieves significant performance gains with negligible training overhead and identical inference latency compared to AA-CLIP.}
%   \small
%   \resizebox{\linewidth}{!}{
%   \begin{tabular}{l|c|cc|cc}
%     \toprule
%     \multirow{2}{*}{\textbf{Method}} & \textbf{Trainable} & \multicolumn{2}{c|}{\textbf{Training Time (Per Epoch)}} & \multicolumn{2}{c}{\textbf{Inference}} \\
%     & \textbf{Params} & \textbf{Stage 1} & \textbf{Stage 2} & \textbf{Latency} & \textbf{FPS} \\
%     \midrule
%     AA-CLIP & 12.6 M & 473.5 s & 614.9 s & 85.7 ms & 11.7 \\
%     \textbf{Proximity-CLIP} & \textbf{22.0 M} & \textbf{465.4 s} & \textbf{640.4 s} & \textbf{85.5 ms} & \textbf{11.7} \\
%     \bottomrule
%   \end{tabular}
%   }
%   \label{tab:efficiency}
% \end{table}

\begin{table}[h]
  \centering
  \caption{Computational efficiency comparison. Proximity-CLIP achieves significant performance gains with negligible training overhead and identical inference latency compared to AA-CLIP.}
  % 使用 footnotesize 减小字体
  \footnotesize 
  \begin{tabular}{l|c|cc|cc}
    \toprule
    \multirow{2}{*}{\textbf{Method}} & \textbf{Trainable} & \multicolumn{2}{c|}{\textbf{Training Time (Per Epoch)}} & \multicolumn{2}{c}{\textbf{Inference}} \\
    & \textbf{Params} & \textbf{Stage 1} & \textbf{Stage 2} & \textbf{Latency} & \textbf{FPS} \\
    \midrule
    AA-CLIP & 12.6 M & 473.5 s & 614.9 s & 85.7 ms & 11.7 \\
    \textbf{Proximity-CLIP} & \textbf{22.0 M} & \textbf{465.4 s} & \textbf{640.4 s} & \textbf{85.5 ms} & \textbf{11.7} \\
    \bottomrule
  \end{tabular}
  \label{tab:efficiency}
\end{table}

During the inference phase, the lightweight AQM module achieves a latency identical to that of AA-CLIP at 11.7 FPS. Consequently, the improvements demonstrated in the main manuscript are achieved with a negligible computational burden.

\section{More Qualitative Results}
\begin{figure}[!t] % 如果需要跨越左右双栏，改为 \begin{figure*}[!t]
    \centering
    \includegraphics[width=\linewidth]{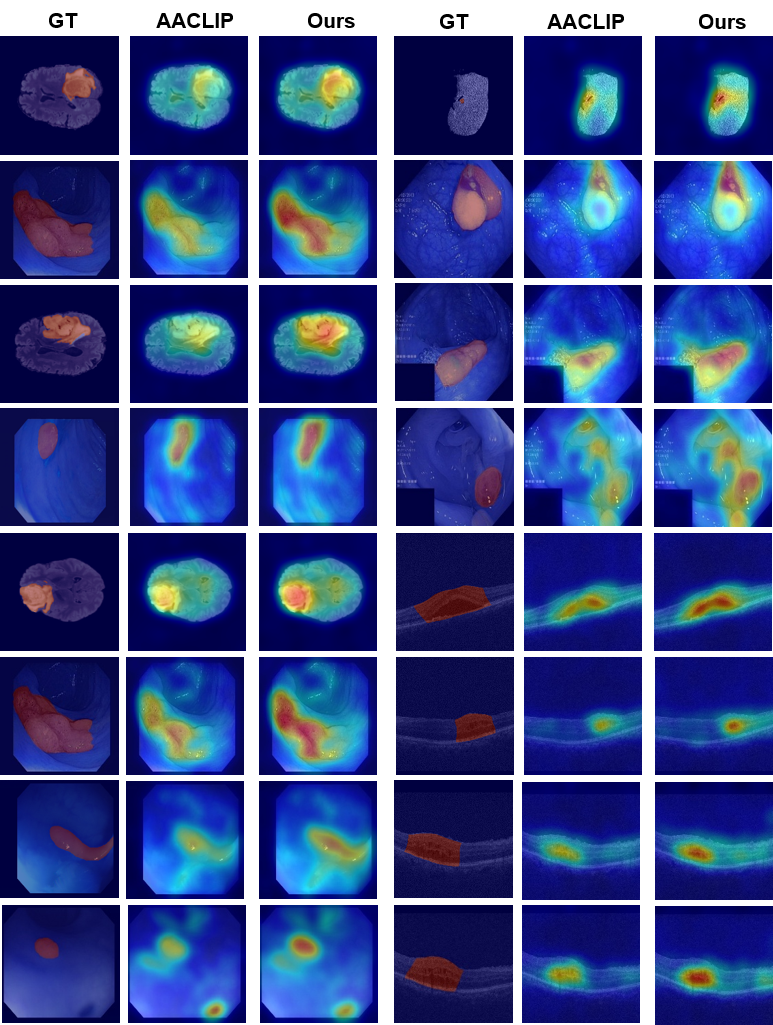}
    \caption{Additional qualitative comparisons of anomaly localization maps between the state-of-the-art method AA-CLIP and our Proximity-CLIP on medical datasets.}
    \label{fig:vis_med_suppl}
\end{figure}

\begin{figure}[!t] % 如果需要跨越左右双栏，改为 \begin{figure*}[!t]
    \centering
    \includegraphics[width=\linewidth]{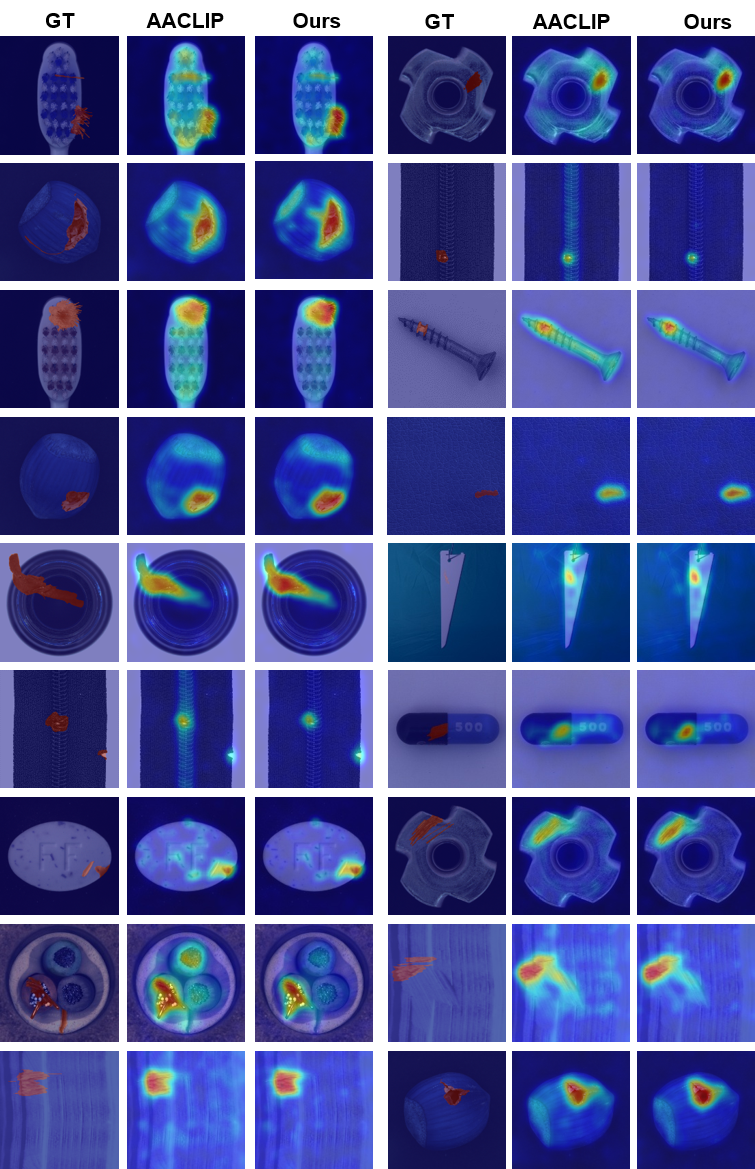}
    \caption{Additional qualitative comparisons of anomaly localization maps between the state-of-the-art method AA-CLIP and our Proximity-CLIP on industrial datasets.}
    \label{fig:vis_ind_suppl}
\end{figure}

We present additional qualitative results in Figure~\ref{fig:vis_med_suppl} and Figure~\ref{fig:vis_ind_suppl} to further demonstrate the superiority of Proximity-CLIP across diverse domains. Specifically, Figure~\ref{fig:vis_med_suppl} illustrates the localization performance on highly complex medical imaging datasets. It is evident that our method consistently isolates subtle anatomical anomalies while effectively suppressing intricate background textures, a challenging scenario where the baseline AA-CLIP frequently suffers from passive feature dilution. Furthermore, Figure~\ref{fig:vis_ind_suppl} provides extended visual comparisons on standard industrial benchmarks. Across various manufacturing defect types, our active semantic spotlight mechanism reliably yields highly concentrated and precise localization masks. These extensive qualitative validations comprehensively substantiate the robustness, accuracy, and cross-domain generalizability of the proposed framework.
% 补充材料中两张可视化图

\end{document}